\documentclass[10pt,twocolumn,letterpaper]{article}

\usepackage[arxiv]{cvpr}      

\usepackage{comment}
\usepackage{arydshln}
\usepackage{multirow}

\definecolor{cvprblue}{rgb}{0.21,0.49,0.74}
\usepackage[pagebackref,breaklinks,colorlinks,allcolors=cvprblue]{hyperref}

\def\paperID{} 
\def\confName{CVPR}
\def\confYear{2026}
\usepackage{amsmath}        
\usepackage{nicefrac}       
\usepackage{microtype}      
\usepackage{xcolor}         
\usepackage{xspace}         
\usepackage{graphicx}
\usepackage{multirow}
\usepackage{wrapfig}
\usepackage{enumitem}

\newcommand{\boldparagraph}[1]{\par\vspace{0.6em}\noindent{\bf #1}}

\def\l1{\ensuremath{\ell_1}\xspace}
\def\l2{\ensuremath{\ell_2}\xspace}

\makeatletter
\DeclareRobustCommand\onedot{\futurelet\@let@token\@onedot}
\def\@onedot{\ifx\@let@token.\else.\null\fi\xspace}

\makeatother
\title{Global-Aware Edge Prioritization for Pose Graph Initialization}
\author{Tong Wei$^1$, Giorgos Tolias$^1$, Ji{\v{r}}{\'\i} Matas$^1$, Daniel Barath$^{2,3}$\\
$^1$ Visual Recognition Group, FEE, Czech Technical University in Prague\\
$^2$ Computer Vision and Geometry Group, ETH Zurich, $^3$ HUN-REN SZTAKI \\
{\tt\small \{weitong, toliageo, matas\}@fel.cvut.cz, \{danielbela.barath\}@inf.ethz.ch}
}

\begin{document}
\maketitle

\begin{abstract}
The pose graph is a core component of Structure-from-Motion (SfM), where images act as nodes and edges encode relative poses. 
Since geometric verification is expensive, SfM pipelines restrict the pose graph to a sparse set of candidate edges, making initialization critical. 
Existing methods rely on image retrieval to connect each image to its $k$ nearest neighbors, treating pairs independently and ignoring global consistency. 
We address this limitation through the concept of edge prioritization, ranking candidate edges by their utility for SfM. 
Our approach has three components: 
(1) a GNN trained with SfM-derived supervision to predict globally consistent edge reliability; 
(2) multi-minimal-spanning-tree-based pose graph construction guided by these ranks; and 
(3) connectivity-aware score modulation that reinforces weak regions and reduces graph diameter. 
This globally informed initialization yields more reliable and compact pose graphs, improving reconstruction accuracy in sparse and high-speed settings and outperforming SOTA retrieval methods on ambiguous scenes. 
The ode and trained models are available at \url{https://github.com/weitong8591/global_edge_prior}.
\end{abstract}

\vspace{0mm}
\section{Introduction}
\vspace{0mm}
Large-scale 3D reconstruction from image collections is a central problem in computer vision.
Structure-from-Motion (SfM) pipelines estimate camera poses and 3D structure from images~\citep{ullman1979interpretation,schonberger2016structure}, enabling applications in virtual reality, visual localization~\citep{panek2023visual}, autonomous driving~\citep{song2014robust}, and novel view synthesis~\citep{song2020deep,Riegler_2021_CVPR,kerbl20233d}.
A fundamental bottleneck shared by SfM pipelines is the construction of an initial \emph{pose graph} -- a sparse set of image pairs selected for geometric verification.
Since verifying all ${n \choose 2}$ pairs is infeasible, the accuracy of this sparse pose graph determines the success and efficiency of the reconstruction.
However, current initialization strategies rely almost exclusively on per-image retrieval, treating pairs independently and failing to integrate global cues, often leading to suboptimal or weakly connected graphs.

SfM pipelines are commonly divided into incremental methods~\citep{schonberger2016structure}, which register images sequentially, and global methods~\citep{pan2024global}, which jointly estimate all poses before bundle adjustment.
Despite their algorithmic differences, both begin with the same steps: pose graph construction~\citep{barath2021efficient}, keypoint detection~\citep{lowe2004distinctive, detone2018superpoint}, feature matching~\citep{sarlin2020superglue, lindenberger2023lightglue}, and geometric verification~\citep{RANSAC, barath2019magsac}.
The pose graph serves as the structural backbone: if important connections are missing at initialization, they are \textit{not} recovered later~\citep{wilson2014robust,manam2024leveraging,damblon2025learning}, limiting downstream accuracy~\citep{zach2010disambiguating, wilson2014robust}.
Thus, building a globally meaningful pose graph \emph{at the very start} is essential.

Traditional initialization connects each image to its $k$ nearest neighbors according to visual descriptors~\citep{arandjelovic2016netvlad, berton2025megaloc}.
This greedy procedure operates independently per image and ignores the global structure.
Moreover, once the initial edges are chosen, later stages \textit{only prune} but do not add connections, thus irreversibly losing the global context.
As a result, strong but non-redundant edges may be overlooked, and the resulting graph may contain elongated chains, poorly conditioned regions, or multiple weakly coupled substructures.

 \begin{figure}[t]
     \centering\includegraphics[width=1.0\linewidth, trim=5 0 0 0, clip]{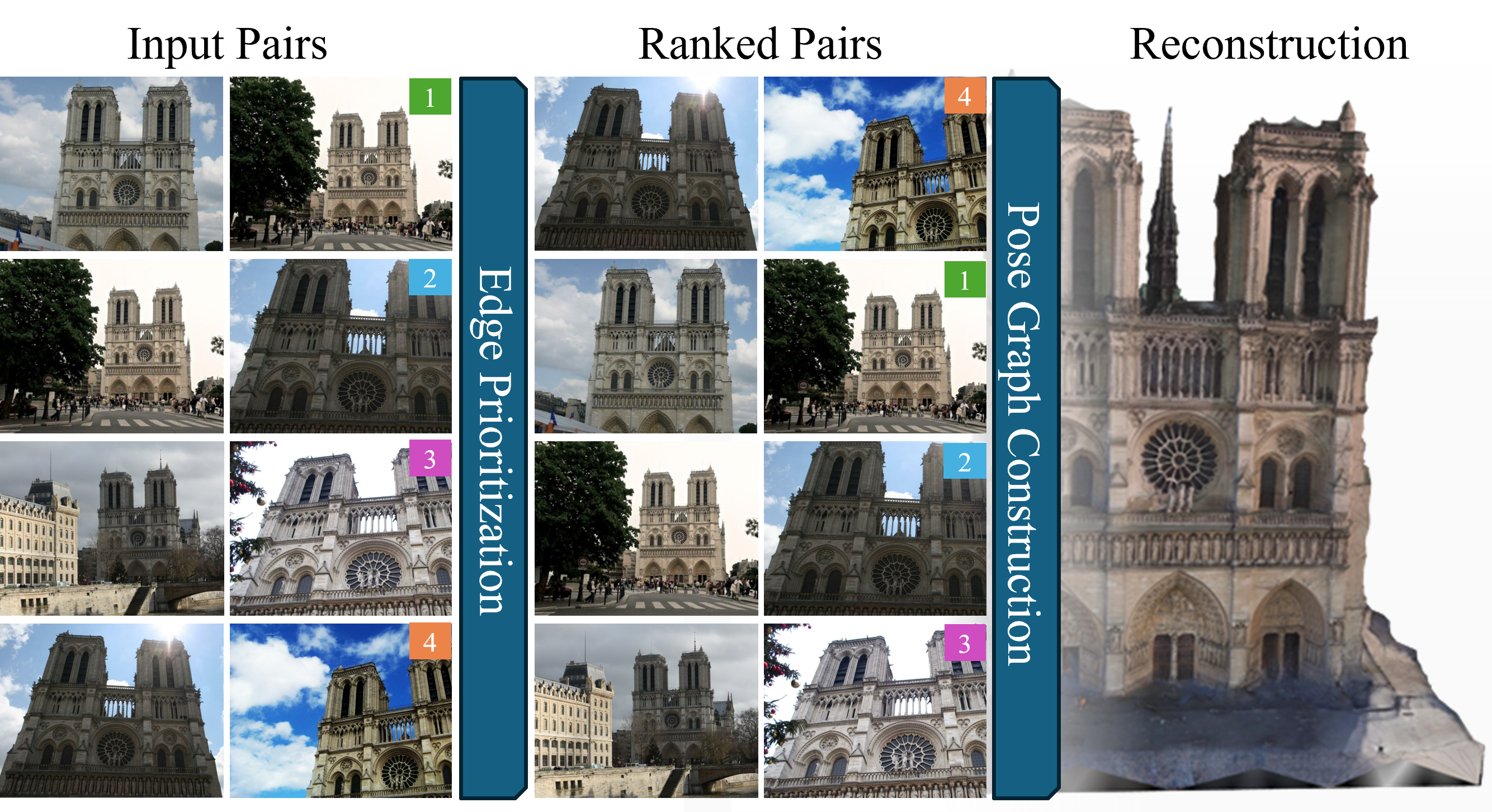}
     \caption{
     Given a set of input image pairs (\textit{left}), our method ranks all candidate edges by global matchability (\textit{middle}) and constructs a compact, well-connected pose graph via multi-MST selection (\textit{right}).  
     The resulting initialization enables accurate and stable 3D reconstruction, even under sparse or ambiguous settings.}
     \label{fig:teaser}
\end{figure}


In this work, we move beyond per-image retrieval and introduce the notion of \emph{edge prioritization}: ranking all candidate edges by their global utility for SfM.
Our method consists of three components.
First, we train a Graph Neural Network (GNN) with SfM-derived supervision to predict globally informed edge scores.
Second, we use these scores to construct the pose graph by iteratively computing multiple minimum spanning trees (MSTs), ensuring global coverage.
Third, we further modulate edge scores across iterations to strengthen weakly connected regions and reduce long relative-pose chains, improving the stability of estimation.
As shown in Fig.~\ref{fig:teaser}, we aim at correct rank and potential image pair retrieval for accurate 3D reconstruction. By integrating global reasoning into both ranking and selection, our method produces significantly more reliable pose graphs, particularly in the sparse and high-speed regime where only a few edges can be verified.

\vspace{0mm}
\section{Related Work}
\vspace{0mm}

\textbf{Structure-from-Motion.}
SfM reconstructs 3D structure and camera poses from image collections~\citep{schonberger2016structure}.  
Incremental pipelines register images sequentially with repeated bundle adjustment; they are robust in practice but can drift and scale poorly.  
Global pipelines~\citep{carlone2015initialization, zhu2018very, cui2015global, pan2024global} instead estimate all poses jointly, typically through motion averaging on an initial pose graph, followed by global bundle adjustment.  
Recent work such as GLOMAP~\citep{pan2024global} demonstrates that global SfM can reach state-of-the-art accuracy at competitive speed.  

Both incremental and global pipelines share the same early-stage steps: pose graph initialization via image retrieval, local feature detection~\citep{lowe2004distinctive, detone2018superpoint}, feature matching~\citep{sarlin2020superglue, lindenberger2023lightglue}, and geometric verification~\citep{RANSAC, barath2019magsac}.  
The initial pose graph is foundational: initialization proposes candidate edges, and verification assigns relative poses and prunes outliers.  
Missing edges are rarely recovered later, and early mistakes propagate into subsequent optimization, making the quality of the initial graph crucial.  

\noindent
\textbf{Pairwise Image Similarity Learning.}
Pose graph initialization is typically driven by image retrieval, where global descriptors map visually similar images to nearby embeddings.  
Classical approaches include VLAD~\citep{jegou2010aggregating} and NetVLAD~\citep{arandjelovic2016netvlad}, while CosPlace~\citep{berton2022rethinking} reframes retrieval as a classification task.  
Recent methods leverage vision foundation models: DINOv2-SALAD~\citep{izquierdo2024optimal} aggregates patch tokens via optimal transport, and MegaLoc~\citep{berton2025megaloc} fine-tunes SALAD to achieve state-of-the-art retrieval accuracy.  

Re-ranking further improves candidate selection. Patch-NetVLAD~\citep{hausler2021patch} enriches descriptors with region-level matching, and VOP~\citep{wei2025breaking} predicts overlap scores for voting-based re-ranking.  
However, all these methods operate strictly pairwise: each potential edge is evaluated in isolation, ignoring global consistency and structural context.  
Given that edges are rarely added after initialization, globally important connections can be lost permanently -- motivating methods that reason over the entire image set.  

\noindent
\textbf{Pose Graph Construction and Filtering.}
Beyond simple retrieval, several works aim to accelerate or refine pose graph construction.  
\citet{barath2021efficient} reuse intermediate pose estimates to speed up graph formation, and \cite{barath2022pose} use Bayesian reasoning to terminate geometric verification early on unmatchable pairs.  
These approaches reduce verification cost but leave the initial choice of candidate edges unchanged.

Most pipelines build the pose graph by connecting each image to its $k$ nearest neighbors based on learned or engineered descriptors~\citep{schonberger2016structure}.  
While effective, $k$NN selection yields primarily local connectivity and may miss globally informative edges.  
Minimum spanning trees (MSTs) offer an appealing alternative: they guarantee global connectivity with minimal edges.  
Prior work has explored single-tree strategies such as sampling MSTs to avoid duplicates~\citep{roberts2011structure}, constructing one MST and adding edges incrementally~\citep{barath2021efficient}, or using a lone MST as a sparse prior in Light3R-SfM~\citep{elflein2025light3r}.  
However, a single tree is structurally fragile -- errors in a few edges can compromise large parts of the graph.

Multiple spanning trees have appeared mainly in \emph{refinement} stages.  
\citet{xiao2021progressive} use the union of MSTs to enforce loop consistency, and \citet{gan2024efficient} build orthogonal trees from engineered similarities to accelerate verification.  
Unlike these works, we construct \emph{multiple} MSTs directly at initialization, guided by learned edge ranks, obtaining complementary connectivity while retaining sparsity.

After initialization, filtering methods remove unreliable edges prior to motion averaging or bundle adjustment.  
\citet{chen2020graph} cluster large graphs for parallel SfM.  
\citet{manam2023sensitivity, manam2024leveraging} detect unstable structures via camera-triplet analysis.  
Doppelganger detection~\citep{cai2023doppelgangers, xiangli2025doppelgangers++} removes visually similar but geometrically inconsistent pairs.  
These methods act only \emph{after} tentative matching and verification -- steps that are computationally expensive and could be avoided with stronger initial selection.

Graph Neural Networks have also been used for pose graph \emph{optimization}.  
PoGO-Net~\citep{li2021pogo} denoises relative poses through message passing. 
\citet{brynte2024learning} use attention-based GNNs for tracking and pose estimation. 
\citet{damblon2025learning} filter outlier edges in an already constructed graph after camera orientation estimation.  
All these methods operate on graphs where relative poses have already been computed.  
In contrast, we predict edge reliability \emph{before} geometric verification and combine it with multi-MST initialization, yielding sparse yet robust pose graphs from the outset.

\vspace{0mm}
\section{Global Edge Prioritization}
\vspace{0mm}

\begin{figure*}
    \centering
    \includegraphics[width=0.93\linewidth]{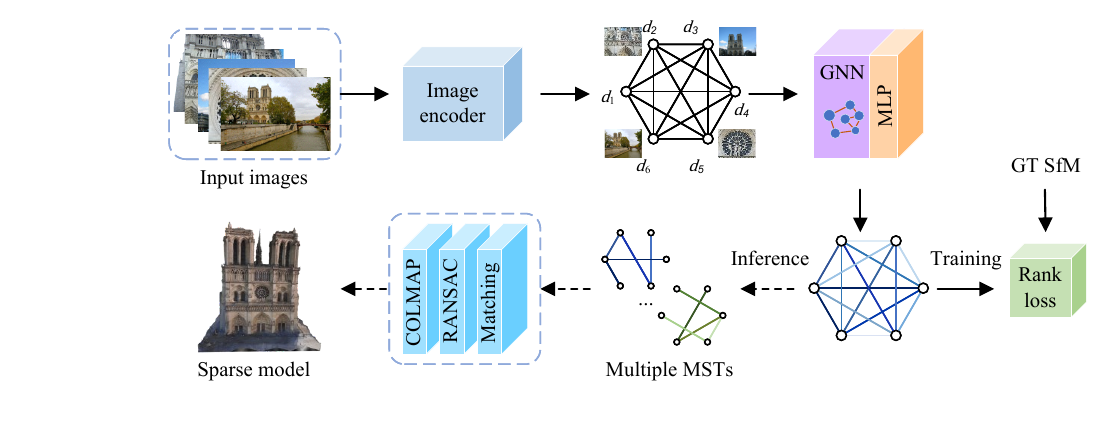}
     \vspace{0mm}
       \caption{
        \textbf{Overall pipeline.} Input images are first encoded using a fine-tuned image encoder (DINOv2 backbone with SALAD aggregation).  
        A complete graph is then constructed over image embeddings and processed by our GNN–MLP model to predict global edge ranks.  
        During training, these predictions are supervised using geometry-derived ranking signals via a differentiable ranking loss.  
        At inference, the predicted ranks guide the construction of multiple minimum spanning trees, whose union forms the initial pose graph.  
        Incremental SfM is finally applied on this graph to recover the sparse 3D reconstruction.}
    \label{fig:pipeline}
\end{figure*}

We formalize pose graph initialization as an \emph{edge ranking} problem and present two core components:  
(i) a GNN-based model trained with geometric supervision to predict global edge ranks, and  
(ii) a connectivity-aware edge selection strategy based on multiple minimum spanning trees (MSTs) combined with distance-based score modulation.  
The full training and inference pipeline is outlined in Fig.~\ref{fig:pipeline}.

\vspace{0mm}
\subsection{Problem Statement}
\vspace{0mm}

Given $N$ input images $I=\{I_{1},\dots,I_{N}\}$, the goal is to construct an \emph{initial pose graph} $\mathcal{G}_0=(\mathcal{V},\mathcal{E}_0)$, where $\mathcal{V}$ indexes the images and $\mathcal{E}_0$ contains the image pairs selected for subsequent geometric verification.  
Since evaluating all ${N\choose 2}$ pairs is infeasible, we assign an edge rank $r_{ij}$ to every pair and select a globally connected edge set:  
\[
\mathcal{E}_0 = \mathrm{Select}(r,\,\text{budget}),
\]
which is likely to yield accurate reconstruction.

Downstream SfM then performs matching and geometric verification \emph{only} on the edges in $\mathcal{E}_0$, producing a verified pose graph $\mathcal{G}_{\text{pose}}=(\mathcal{V},\mathcal{E})$ with $\mathcal{E}\subseteq\mathcal{E}_0$.  
Our contributions target the ranking and selection that define $\mathcal{G}_0$; the remainder of the pipeline remains unchanged.

\vspace{0mm}
\subsection{Edge Ranking Prediction}
\vspace{0mm}

\boldparagraph{Image Encoder.}
Each image $I_i$ can be encoded into a descriptor $d_i = f_\text{en}(I_i) \in \mathbb{R}^{d}$.  
A naive rank estimator would use cosine similarity $\langle d_i, d_j\rangle$, but such pairwise scores ignore the structure of the full image collection.  
We therefore allow global reasoning via message passing in GNN. 

\boldparagraph{Graph Neural Network.}
To incorporate global reasoning beyond pairwise similarity, we construct a complete graph over image embeddings $\{d_i\}$.  
The nodes in the input graphs are consistent as in the pose graph, while the edges are two-view relative motions. Thus, our goal is to rank all the edges in a global manner by their potential to be connected.
Following~\citep{turkoglu2021visual}, each directed edge $(i,j)$ is initialized with a feature vector formed from the embeddings of its endpoints and their cosine similarity via a fully connected layer $f_l$ with ReLU activation:
\[
e_{ij}^{\,0} = \mathrm{ReLU}\!\left(f_l\big[d_i,\, d_j,\, \langle d_i,d_j\rangle\big]\right),
\qquad e_{ij}^{\,0}\in\mathbb{R}^{d_l}.
\]
We apply two iterations of edge-node message passing to propagate information across the full image set.  
In iteration $t$, edge features are first updated using the current node embeddings as follows:
\[
e_{ij}^{\,t} = f_{\mathrm{edge}}\!\left([e_{ij}^{\,t-1},\, d_i^{\,t},\, d_j^{\,t}]\right),
\]
where $f_{\mathrm{edge}}$ is a two-layer MLP with batch normalization and ReLU activations.

Next, each node gathers messages from all its neighbors.  
A message from node $j$ to node $i$ is computed as:
\[
m_{ji}^{\,t} = f_{\mathrm{msg}}\!\left([e_{ij}^{\,t},\, d_j^{\,t}]\right),
\]
where $f_{\mathrm{msg}}$ shares the same architecture as $f_{\mathrm{edge}}$ but with different input dimensionality.  
Messages are averaged over all neighbors as follows:
\[
m_i^{\,t} = \frac{1}{N}\sum_{j=1}^{N} m_{ji}^{\,t}.
\]
Node embeddings are updated via another two-layer MLP:
\[
d_i^{\,t+1} = f_{\mathrm{update}}\!\left([d_i^{\,t},\, m_i^{\,t}]\right).
\]
This process is repeated twice, enabling each edge to aggregate information that depends not only on its endpoints but also on their global context in the full image set.  

Finally, the refined edge features from the second iteration are passed through a prediction MLP (two linear layers with ReLU and dropout) to obtain the final edge rank as:
\[
\hat{r}_{ij} = f_{\mathrm{MLP}}\!\left(e_{ij}^{\,2}\right).
\]
This allows the model to integrate information across all image pairs, enabling edge ranking that reflects global structural relations rather than solely local embedding similarity.

\boldparagraph{Geometry-based Supervision.}
To align ranking with the actual requirements of SfM, we supervise the model using signals derived directly from 3D reconstruction.  
Unlike standard retrieval labels, which reflect only visual similarity, these signals capture whether a pair is truly useful for recovering accurate camera poses.

Given a set of training images, we first perform keypoint detection, feature matching, and relative pose estimation with RANSAC.  
For each candidate pair $(i,j)$, we record the number of inlier correspondences returned by RANSAC:
\begin{equation}
    u_{ij} = \#\{\text{RANSAC inliers for } (i,j)\}.
\end{equation}
Large values of $u_{ij}$ correlate with stable two-view geometry, while low values indicate either limited visual overlap or incorrect matches.  
Although RANSAC is robust, its estimates can still be noisy or occasionally wrong; therefore we additionally filter pairs with low inlier counts before feeding them into the SfM pipeline.

To complement this signal, we also compute for each pair the number of jointly seen triangulated points as:
\begin{equation}
    v_{ij} = \#\{\text{3D points visible in both } I_i \text{ and } I_j\},
\end{equation}
which reflects how well the two views contribute to consistent multi-view geometry.  
Pairs with high $v_{ij}$ typically correspond to stable baselines and reliable camera constraints, even if their initial RANSAC inlier counts were modest.

The two signals capture complementary aspects of geometric utility:  
$u_{ij}$ measures the immediate verifiability of the pair, whereas $v_{ij}$ reflects its long-term usefulness once a global reconstruction is formed.  
To obtain ground-truth edge ranks, both signals are normalized to $[0,1]$ and combined via
\begin{equation}\label{eq:score}
    \tilde{r}_{ij} = \tfrac{1}{2}\left(\text{norm}(u_{ij}) + \text{norm}(v_{ij})\right),
\end{equation}
where normalization spreads values below $1000$ inliers over $[0.0,0.8]$, and larger counts over $[0.8,1.0]$, ensuring that easy pairs do not dominate the training objective.

This supervision is entirely \emph{self-supervised}: $u_{ij}$ and $v_{ij}$ are obtained automatically from standard SfM pipelines without human annotation.  
At inference time, no geometric verification runs. 
The model predicts edge ranks solely from image embeddings and global context provided by the GNN.

\boldparagraph{Loss Function.}
In contrast to works that train image encoders with categorical or binary labels~\citep{berton2025megaloc,izquierdo2024optimal}, our supervision is continuous.  
We care about the \emph{relative} ordering of image pairs, not the exact value of their labels.  
Thus, we train the model as a ranking rather than a regression problem.  

The quality of a predicted ranking is commonly evaluated in information retrieval with Normalized Discounted Cumulative Gain (NDCG)~\citep{jarvelin2002cumulated}.  
DCG measures the quality of a ranked list by summing item relevance, discounted for lower-ranked items.  
For ground-truth ranks $r_i \in \{1, \dots, M\}$ ($r_i=1$ for top-1), the relevance score of item $i$ is $v_i = M - r_i$.  
Given predicted edge ranks, pairs are sorted, and for each item $i$ with predicted rank $\hat{r}_i$, DCG is computed as:  
\begin{equation}\label{eqn:dcg}
    \text{DCG} = \sum_{i=1}^{M} \frac{2^{v_i} - 1}{\log_2(\hat{r}_i + 1)}.
\end{equation}
The ideal DCG (IDCG) is computed from the ground-truth ranking, and NDCG is defined as $\text{NDCG} = \text{DCG} / \text{IDCG}$, ranging from zero to one.  

Since NDCG is non-differentiable, we adopt NDCGLoss2++~\citep{wang2018lambdaloss}, built upon the LambdaRank algorithm~\citep{burges2006learning, burges2010ranknet}, which optimizes a differentiable approximation of NDCG by considering pairwise item swaps and their effect on ranking quality.  
Its detailed formulation is provided in~\citep{wang2018lambdaloss}.
This loss has been shown effective for training MLP-based ranking models~\citep{pobrotyn2020context}.  
The only hyperparameter is $k$, which determines the number of items sampled per list, set to half the list size in our training.

\vspace{0mm}
\section{Connectivity-Aware Score Modulation}
\vspace{0mm}

\boldparagraph{Selection with Minimum Spanning Trees (MSTs).}
Given predicted edge ranks $\hat{r}_{ij}$, the goal is to select a sparse subset of candidate pairs that is geometrically meaningful for SfM.  
The standard choice in most pipelines is $k$-nearest neighbor selection, where each image connects to its top-$k$ neighbors based on embedding similarity~\citep{schonberger2016structure,pan2024global}.  
However, $k$NN selection is local -- it examines each image independently and does not account for how the resulting edges jointly shape the global topology of the pose graph.  
In particular, $k$NN construction can yield fragmented or weakly connected structures, creating long chains or disconnected components that significantly degrade global pose estimation.

To enforce global connectivity in a principled way, we adopt Minimum Spanning Trees (MSTs).  
An MST is the subset of $N-1$ edges that connects all images with minimum total weight.  
We define edge weights as:
\[
w_{ij} = 1 - \hat{r}_{ij},
\]
so high-confidence edges are preferentially selected by MST construction.  
The resulting tree is globally connected and sparse, making it an attractive backbone for SfM initialization.  
Each MST is found by Kruskal’s algorithm~\citep{kruskal1956shortest}.

\boldparagraph{Multiple MSTs for Redundancy.}
Although a single MST guarantees connectivity, it is brittle: each pair of nodes is connected by exactly one path.  
This structure often yields long chains of cameras whose constraints propagate through many intermediaries, increasing sensitivity to noise and drift.  
A single MST is also vulnerable to local errors: if one high-ranked edge is incorrect or discarded during geometric verification, large parts of the graph become disconnected.

To mitigate this, we construct not one but $k$ MSTs.  
Given the complete graph and predicted ranks, we proceed as:
\begin{enumerate}[label=(\arabic*)]
    \item Compute the first tree as $\mathcal{T}_1 = \mathrm{MST}(\{w_{ij} = 1 - \hat{r}_{ij}\})$.
    \item For $m > 1$, penalize edges already selected in $\mathcal{T}_1, \dots, \mathcal{T}_{m-1}$ by assigning $\infty$ cost, then compute $\mathcal{T}_m$.
    \item Form the initial pose graph as the union of all trees, $\mathcal{G}_\text{init} = \bigcup_{m=1}^k \mathcal{T}_m$.
\end{enumerate}
This ensures that every camera is connected through at least $k$ independent paths, providing structural redundancy even before geometric verification.  
However, a key observation is that multiple MSTs alone \emph{do not} ensure that the graph is well-connected.  
Because MSTs select edges solely based on the local quantity $\hat{r}_{ij}$, they may repeatedly favor edges within dense clusters or along the same dominant chain, leaving the global diameter of the graph large.  
Thus, the graph may contain several disjoint “weak links’’ or elongated regions where poses propagate across many hops.  

These limitations motivate a principled strategy that incorporates \emph{global connectivity} into the selection process.

\boldparagraph{Global Connectivity Representation.}
To explicitly reason about the connectivity quality of the graph as it is constructed, we introduce a representation based on shortest-path distances.  
After selecting MSTs $\mathcal{T}_1,\dots,\mathcal{T}_{m-1}$, we define the current graph as $\mathcal{G}^{(m-1)} = \bigcup_{\ell=1}^{m-1}\mathcal{T}_\ell$.
Although $\mathcal{G}^{(m-1)}$ is connected (starting from $m=2$), its internal structure may still contain long chains or loosely coupled clusters.
For each image pair $(i,j)$, we compute the hop-count shortest-path distance:
$d^{(m-1)}(i,j)$,
with $d^{(m-1)}(i,j)=+\infty$ if the nodes remain disconnected (possible at $m=1$).  
We also compute the graph diameter as:
\[
D^{(m-1)} = \max_{i,j} d^{(m-1)}(i,j).
\]
To normalize distances:
\begin{equation}
    \bar d^{(m-1)}(i,j) =
    \begin{cases}
        \dfrac{d^{(m-1)}(i,j)}{D^{(m-1)}} & D^{(m-1)} < +\infty,\\[1.2ex]
        1 & \text{otherwise}.
    \end{cases}
\end{equation}
Thus, $\bar d^{(m-1)}(i,j)$ is a global context signal: it is high for pairs that are far apart in the current graph.

\boldparagraph{Distance-modulated Score.}
To address that the raw ranks $\hat{r}_{ij}$ do not measure global connectivity, we introduce a mechanism that modulates the predicted ranks using the graph distances.  
For MST iteration $m$, we compute:
\begin{equation}
    s^{(m)}_{ij}
    = (1 - \lambda) \, \hat{r}_{ij}
    + \lambda\, \bar d^{(m-1)}(i,j),
    \label{eq:score_update}
\end{equation}
with modulation weight $\lambda\in[0,1]$.
This formulation has two key effects:
(\textit{1}) If $(i,j)$ is already well-connected in $\mathcal{G}^{(m-1)}$, then $\bar d^{(m-1)}(i,j)$ is small and the score is dominated by $\hat{r}_{ij}$.  
(\textit{2}) If $(i,j)$ lies across a weakly linked region, $\bar d^{(m-1)}(i,j)$ is large, boosting making it more likely to be selected.

The result is a refined ranking that favors edges which both are locally matchable and globally strengthen the topology of the pose graph by reducing its diameter.

After constructing each MST, we update the scores only for the top five candidate edges per image (based on predicted rank) and discard all edges with predicted rank below $0.9$.  
This prevents unreliable pairs from receiving boosts.  
Finally, edges already selected in previous MSTs are masked out by assigning $-\infty$ to their scores before the next iteration.

\boldparagraph{Notes.}
Before the first tree ($m=1$), $\mathcal{G}^{(0)}$ is empty, and thus $\bar{d}^{(0)}(i,j)=1$ for all pairs.  
Eq.~\eqref{eq:score_update} then reduces to a uniform rescaling, so the first MST depends only on $\hat{r}_{ij}$.  
For $m\ge2$, the modulation becomes meaningful, prioritizing distant yet strong pairs that connect separate subgraphs. 

\boldparagraph{Graph Clustering.}
To scale to large image collections, we employ graph clustering~\citep{chen2020graph, damblon2025learning}.  
The complete image graph is partitioned into subgraphs before being processed by our GNN model for edge rank prediction.  
This partitioning prevents GPU memory exhaustion when handling large graphs.  
Following \citet{damblon2025learning}, we use METIS~\citep{karypis1997metis} for graph partitioning, guided by similarities derived from intermediate features of our encoder before the GNN.  
Predictions from different subgraphs are then aggregated to recover the final edge rank matrix for the entire dataset.






\vspace{0mm}
\section{Experiments}
\vspace{0mm}

\begin{figure*}[t]
    \centering
    \vspace{-1mm}
    
    \begin{minipage}{0.315\textwidth}
        \centering
        \includegraphics[width=\textwidth, trim=5 0 0 0, clip]{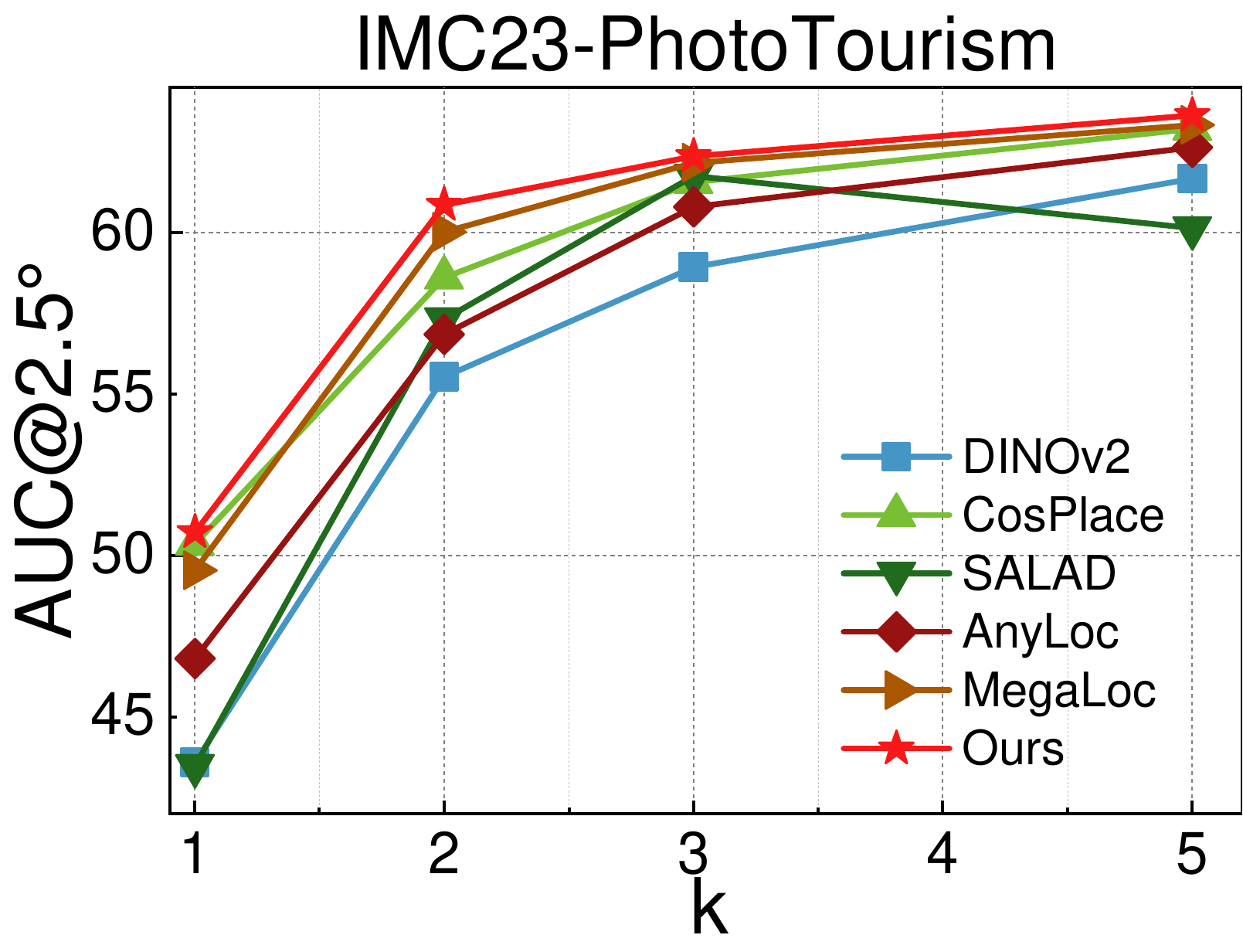}
    \end{minipage}
    \hfill
    \begin{minipage}{0.315\textwidth}
        \centering
        \includegraphics[width=\textwidth, trim=5 0 0 0, clip]{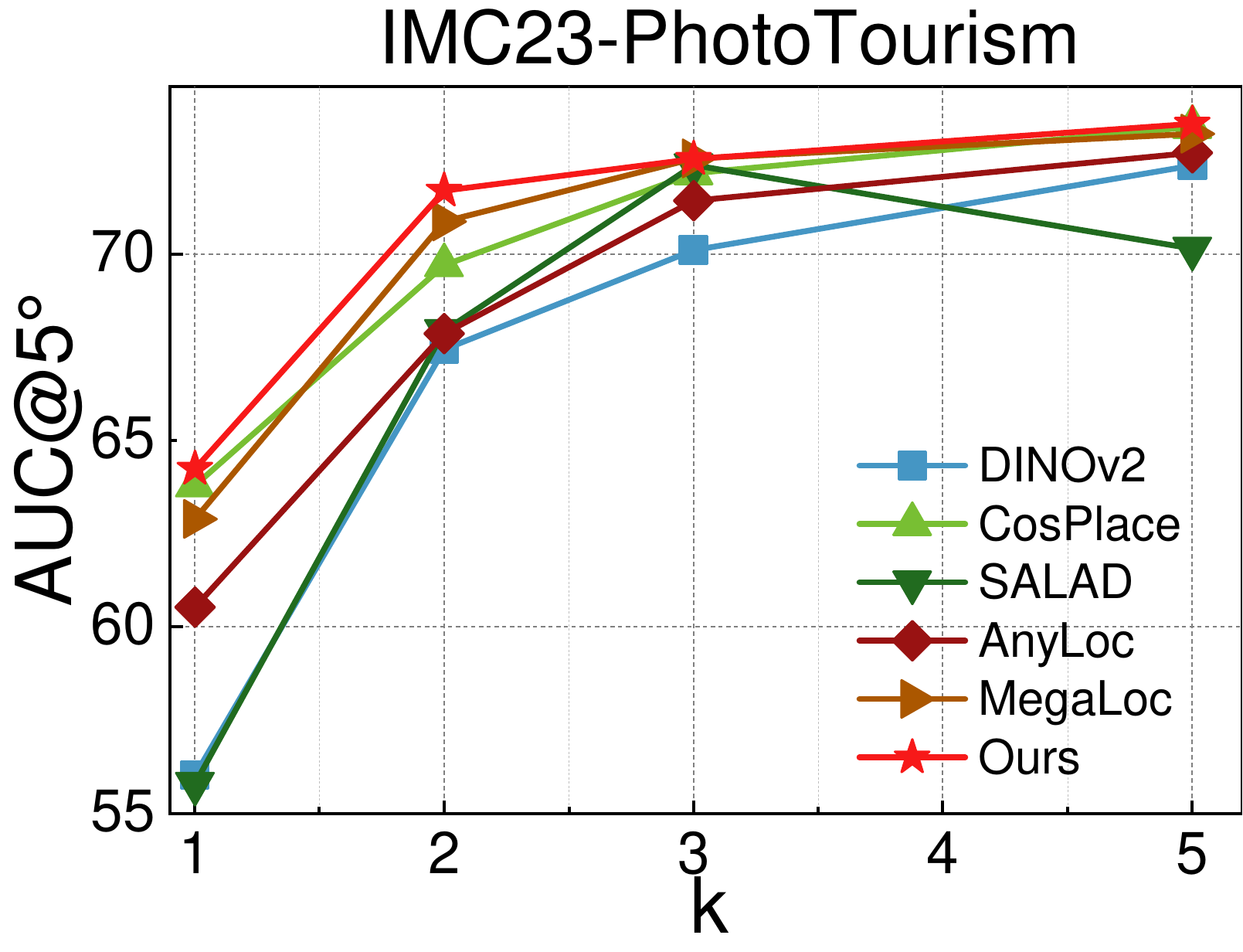}
    \end{minipage}
        \hfill
            \begin{minipage}{0.315\textwidth}
        \centering
        \includegraphics[width=\textwidth, trim=5 0 0 0, clip]{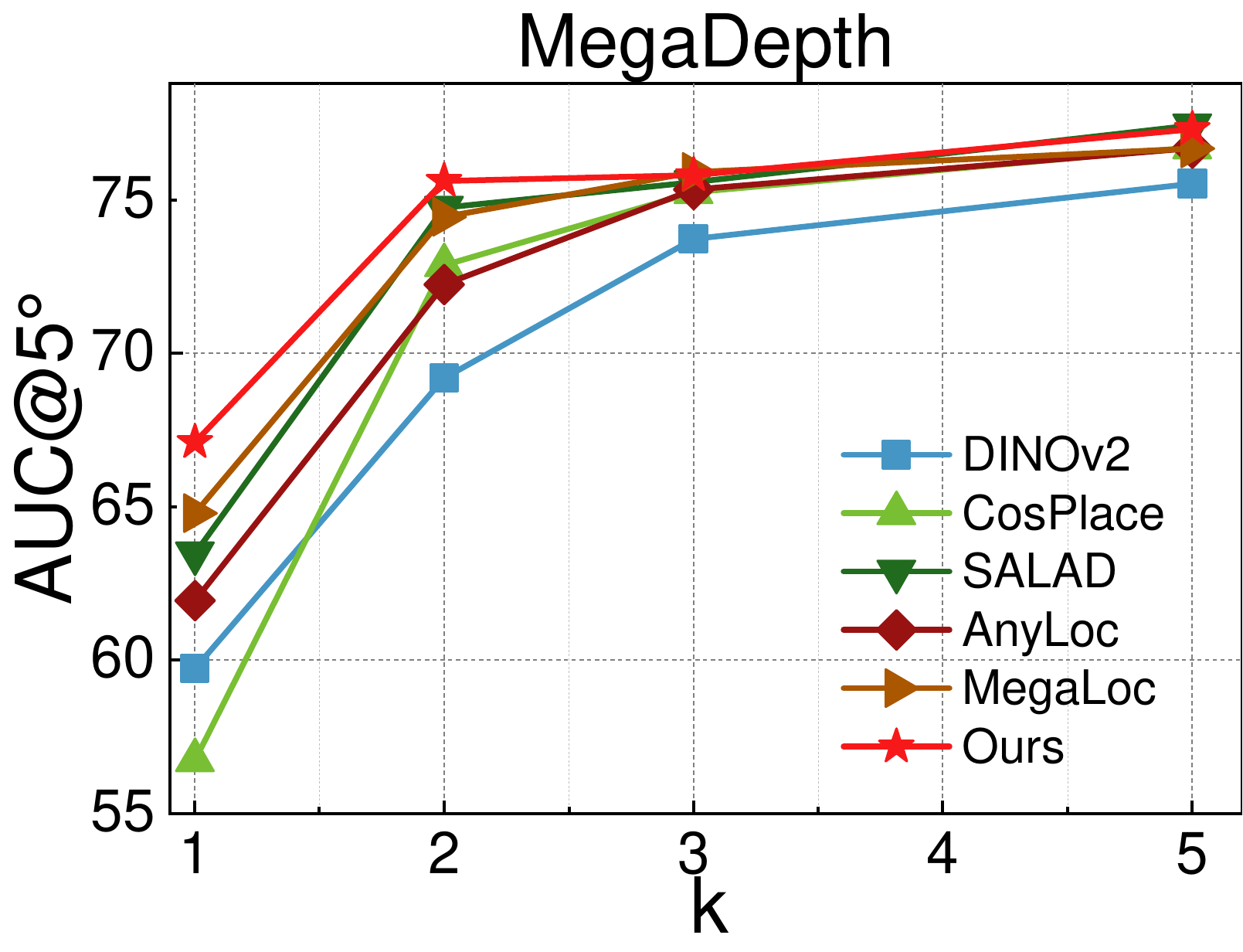} 
    \end{minipage}
    \hfill

\vspace{0.2cm}
    \begin{minipage}{0.315\textwidth}
        \centering
        \includegraphics[width=\textwidth, trim=5 0 0 0, clip]{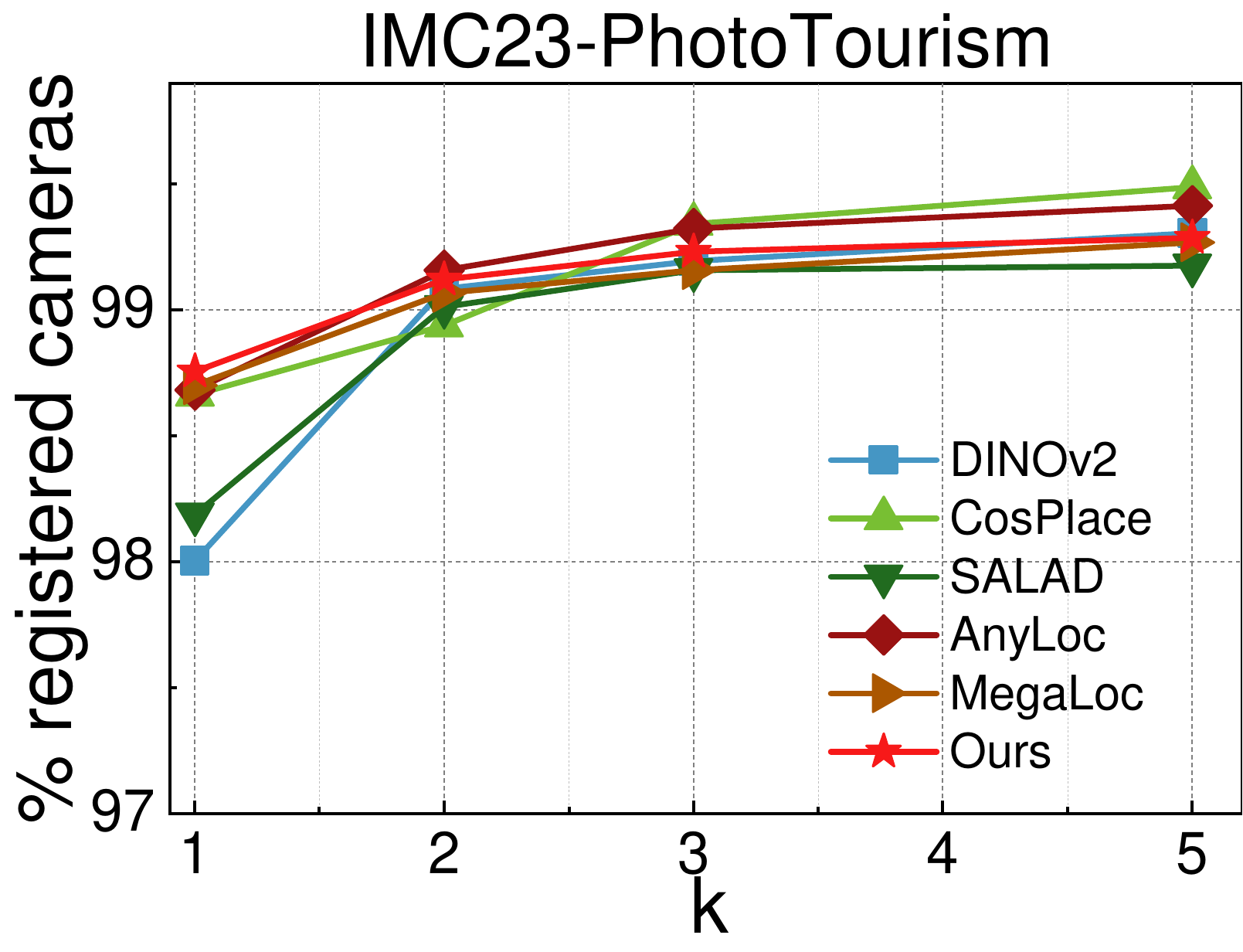}
    \end{minipage}
    \hfill
    \begin{minipage}{0.315\textwidth}
        \centering
        \includegraphics[width=\textwidth, trim=5 0 0 0, clip]{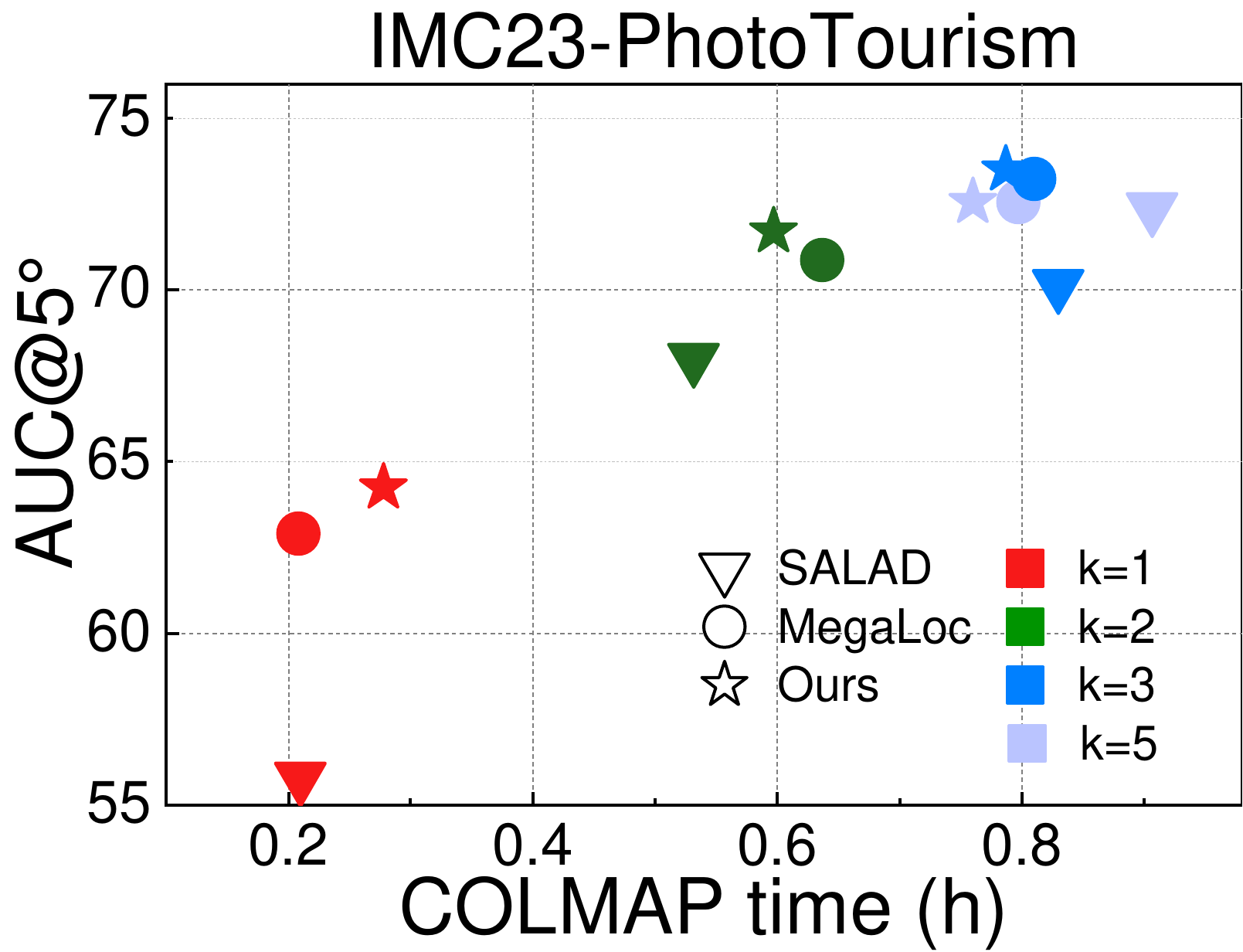}
     \end{minipage}\hfill
    \begin{minipage}{0.315\textwidth}
        \centering
\includegraphics[width=\textwidth, trim=5 0 0 0, clip]{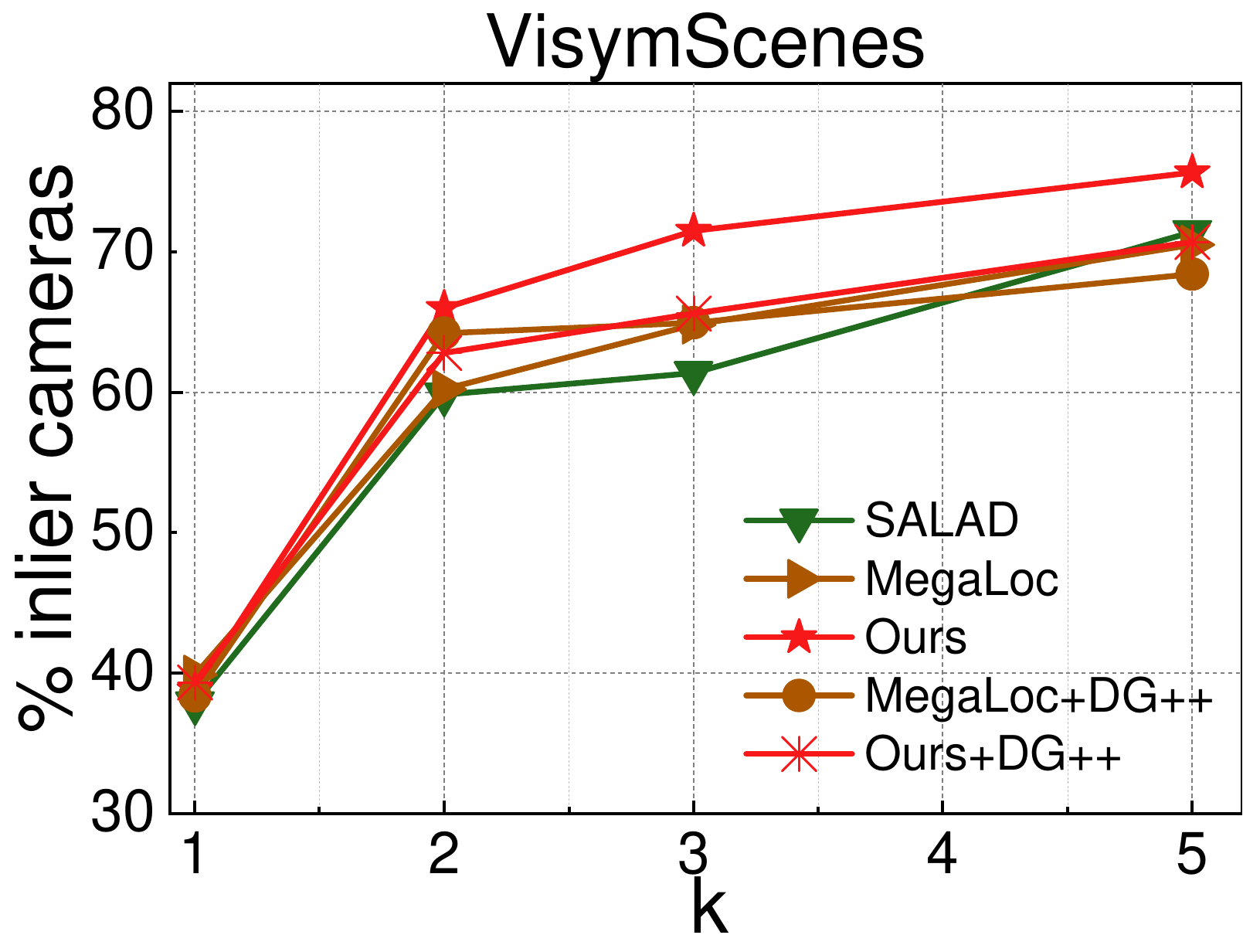}
    \end{minipage}
    \caption{
        COLMAP reconstruction~\cite{schonberger2016structure} performance using pose graphs constructed from multiple MSTs guided by baseline embedding similarities or our learned global edge ranks.  
        \textbf{Top row:} Relative pose accuracy on IMC23-PhotoTourism~\cite{imc2023} (AUC@2.5$^\circ$, left; AUC@5$^\circ$, middle) and on MegaDepth~\cite{li2018megadepth} (AUC@5$^\circ$, right) as the number of MSTs $k$ increases.  
        \textbf{Bottom row:} Percentage of registered cameras on PhotoTourism (left), AUC@5$^\circ$ versus COLMAP runtime on PhotoTourism (middle), and percentage of accurately reconstructed cameras on VisymScenes~\cite{xiangli2025doppelgangers++} (right).  
        Across all benchmarks, our method consistently achieves the highest accuracy, particularly in the sparse regime ($k=1$–$2$), while maintaining competitive or superior reconstruction efficiency.
    }
        
    \label{fig:main}
    \vspace{0mm}
\end{figure*}

We evaluate our approach on large-scale SfM benchmarks, focusing on the quality of the initial pose graphs and their impact on final COLMAP reconstructions.  
Throughout, we compare pose graphs built from multiple MSTs guided either by off-the-shelf image embeddings or by our learned edge ranks.  
For all baselines, descriptor similarities are converted into symmetric similarity matrices, so that all methods operate in the same ``edge-score'' format as our predictor.

\boldparagraph{Datasets.}
Our model is trained on the training split of MegaDepth~\citep{li2018megadepth}, where we construct geometry-based ground-truth edge ranks as described in \eqref{eq:score}.  
For evaluation, we use held-out MegaDepth scenes and 15 PhotoTourism scenes from the Image Matching Challenge 2023 (IMC23)~\citep{imc2023}.  
To assess robustness under strong visual ambiguity, we additionally evaluate on VisymScenes~\citep{xiangli2025doppelgangers++}, which contains doppelganger images with repeated facades and strong structural symmetries.

\boldparagraph{Backbone and Features.}
We use MegaLoc~\citep{berton2025megaloc} as image encoder, combining frozen DINOv2 features with SALAD aggregation and linear projections.  
Our GNN-based edge rank predictor is trained on top of these descriptors.  
For COLMAP, we use SuperPoint~\citep{detone2018superpoint} and LightGlue~\citep{lindenberger2023lightglue} on MegaDepth and IMC23.  
On VisymScenes, following~\citep{xiangli2025doppelgangers++}, we use SIFT~\citep{lowe2004distinctive} with brute-force matching~\citep{bradski2000opencv}.

\boldparagraph{Baselines.}
We compare to state-of-the-art global retrieval and similarity methods: CosPlace~\citep{berton2022rethinking}, AnyLoc~\citep{keetha2023anyloc}, DINOv2-SALAD~\citep{izquierdo2024optimal}, and MegaLoc~\citep{berton2025megaloc}, as well as frozen DINOv2 features.  
For each method, we construct MST-based pose graphs from their similarity scores and feed the resulting graphs into the default COLMAP pipeline.  
We do not use ground-truth intrinsics. 
COLMAP estimates intrinsics as in realistic SfM deployments.

\boldparagraph{Training Details.}
We train on 153 MegaDepth scenes and hold out 8 scenes for validation.  
Each batch contains $240$ images from a single scene, with at most four batches per scene.  
We compute the ranking loss on the top half of edges according to $\tilde r_{ij}$.  
Training runs for 50 epochs with AdamW~\citep{loshchilov2019decoupled} (learning rate $10^{-5}$) on NVIDIA Tesla H200 GPUs (141GB, CUDA 12.8).  
COLMAP is executed on CPU-only machines.

\boldparagraph{Evaluation Protocol.}
We evaluate reconstructions using relative pose accuracy between \emph{all} estimated cameras.  
Following~\citep{pan2024global}, we report Area Under the Recall Curve (AUC) at $2.5^\circ$ and $5^\circ$, the percentage of successfully registered cameras, and COLMAP mapping time.  
Unless stated otherwise, results are shown for pose graphs obtained from $k\in\{1,2,3,5\}$ MSTs.

On VisymScenes, we reconstruct four test scenes using $k\in\{1,2,3,5\}$ MSTs.  
We fix RANSAC to 10k iterations and use a fixed random seed for repeatability.  
Following~\citep{xiangli2025doppelgangers++}, we evaluate the percentage of cameras localized within a fixed geolocation threshold.  
Unlike DG++, which normalizes inlier counts relative only to registered cameras, we normalize by the total number of geo-tagged images, penalizing unregistered views and obtaining a stricter metric.

\boldparagraph{Scalability via Graph Clustering.}
For large collections ($N>500$), we partition the complete graph into subgraphs using METIS~\citep{karypis1997metis}, following~\citep{damblon2025learning}.  
The number of clusters is set to $n_{\text{clusters}} = 1 + \lfloor N / N_{\text{max}}\rfloor$.  
Each subgraph is expanded with its 1-hop neighbors (without duplication) and processed by the GNN. 
Predictions for overlapping pairs are averaged to form the final global edge ranking matrix.

We now analyze 3D reconstruction performance on IMC23-PhotoTourism, MegaDepth, and VisymScenes, followed by ablations on edge selection, connectivity-aware score modulation, and training choices.

\vspace{0mm}
\subsection{3D Reconstruction}
\vspace{0mm}

\boldparagraph{IMC23-PhotoTourism.}
Figure~\ref{fig:main}~(top left and middle) shows AUC@2.5$^\circ$ and AUC@5$^\circ$ as a function of $k$.  
Across all values of $k$, our approach yields the highest pose accuracy.  
The margins are largest at $k=1$ and $k=2$, where pose graphs are extremely sparse and global reasoning is most important.  
As $k$ increases, all strong baselines gradually converge toward similar high-accuracy solutions, yet our method consistently remains on top.

An important observation is that all baselines perform noticeably better under the multi-MST selection framework than under their native $k$NN-based selection commonly used in SfM pipelines.  
MST construction provides a stronger structural prior, benefiting all methods.  
A direct comparison against $k$NN selection is provided later in the ablation study.

Figure~\ref{fig:main}~(bottom left) shows the percentage of registered cameras.  
All methods achieve nearly complete reconstructions, clustering around $99\%$ across all values of $k$.  
Thus, improvements in AUC stem from selecting more informative edges rather than registering more cameras.
Figure~\ref{fig:main}~(bottom middle) reports AUC@5$^\circ$ versus COLMAP runtime.  
Across all $k$, our method lies on the Pareto frontier: for comparable runtimes, it consistently achieves higher accuracy than competing methods.

\boldparagraph{MegaDepth.}
The top-right plot of Figure~\ref{fig:main} reports AUC@5$^\circ$ on the MegaDepth test scenes ``0015'' and ``0022'' as a function of $k$.  
As on IMC23, our curve is consistently the best across all numbers of MSTs.  
The gains are again largest at $k=1$ and $k=2$, where retrieval-based methods still suffer from missing or suboptimal long-range edges.  
For larger $k$, all strong baselines converge toward similar high accuracy, but our method maintains a small lead at every point.  
These results mirror the PhotoTourism findings and indicate that global edge prioritization is particularly beneficial when the pose graph must remain very sparse, while remaining competitive in the denser regime.

\vspace{0mm}
\subsection{Disambiguating SfM}
\vspace{0mm}

We further evaluate our method under severe visual ambiguity using VisymScenes~\citep{xiangli2025doppelgangers++}, a dataset containing many \emph{doppelganger} images -- visually similar yet geometrically unrelated views.  
Each of the four test scenes mixes genuine overlapping images with hard negative distractors, often leading to multiple disconnected components in COLMAP reconstructions.
For initialization, we construct MST-based pose graphs using SALAD similarity, MegaLoc similarity, or our predicted global edge ranks.  
We run COLMAP and report the percentage of cameras whose poses fall within a fixed geolocation threshold.
We use the model trained on MegaDepth for all the tests, no retraining needed.

Figure~\ref{fig:main}~(bottom right) and Table~\ref{tab:visym} show the results.  
Across all $k$, our method reconstructs the largest fraction of correct cameras.  
Even though the baselines improve with increasing $k$, they remain consistently below our method.  
At $k=5$, our method exceeds $75\%$ correct reconstructions, clearly outperforming SALAD and MegaLoc.  
Applying DoppelGanger++ on top of our scores provides no further benefit, indicating that our predictor inherently suppresses misleading edges.
These results demonstrate that global edge prioritization is highly effective in ambiguous settings where local visual similarity fails to distinguish distractors.  
Despite the distribution shift from MegaDepth and PhotoTourism, our model generalizes well without retraining.

\begin{table}[t]
\vspace{0mm}
\centering
\centering
\resizebox{0.46\textwidth}{!}{%
\begin{tabular}{lcccc| c}
\hline
Method & Config & \multicolumn{3}{c}{AUC@5$^\circ~\uparrow$} & Time (min) $\downarrow$ \\
$k$ MSTs & $\rightarrow$ &  2 & 3 & 5 & 5 \\
\hline
\hline
 \multirow{3}{*}{MegaLoc}    & w/o modul.       & 60.2 & 64.8 & 70.5 & \phantom{1}7.3 \\
    & w/\phantom{o} modul.        & {64.1} & \textbf{65.3} & \textbf{74.7} & 14.1 \\
    & DG++        &  \textbf{64.2} & 64.9 & 68.4 & \phantom{1}8.6 \\
\hline
\multirow{3}{*}{Ours}       &   w/o modul.       & {61.9} & 65.8 & 71.7 & \phantom{1}5.7 \\
 & w/\phantom{o} modul.        & \textbf{66.0} & \textbf{71.5} & \textbf{75.6} & 13.3 \\
  & DG++        & {62.8} & {65.6} & {70.7} & \phantom{1}7.4 \\
\hline
\end{tabular}%
}
\vspace{0em}
\caption{\textbf{Ablation} of connectivity-aware score modulation on VisymScenes~\cite{xiangli2025doppelgangers++}.  
We report AUC@5$^\circ$ for pose graphs built from MegaLoc similarities and from our predicted global edge ranks, with and without modulation, as well as with DoppelGanger++ filtering.  
Modulation consistently improves accuracy for both methods, especially in the sparse regime ($k=2$--$3$).  
COLMAP mapping time at $k=5$ is shown in minutes.}
\label{tab:visym}
\end{table}

\vspace{0mm}
\subsection{Ablations}
\vspace{0mm}

\begin{figure*}[t]
    \centering
    \vspace{0mm}
  \includegraphics[width=0.33\textwidth, trim=5 0 0 0, clip]{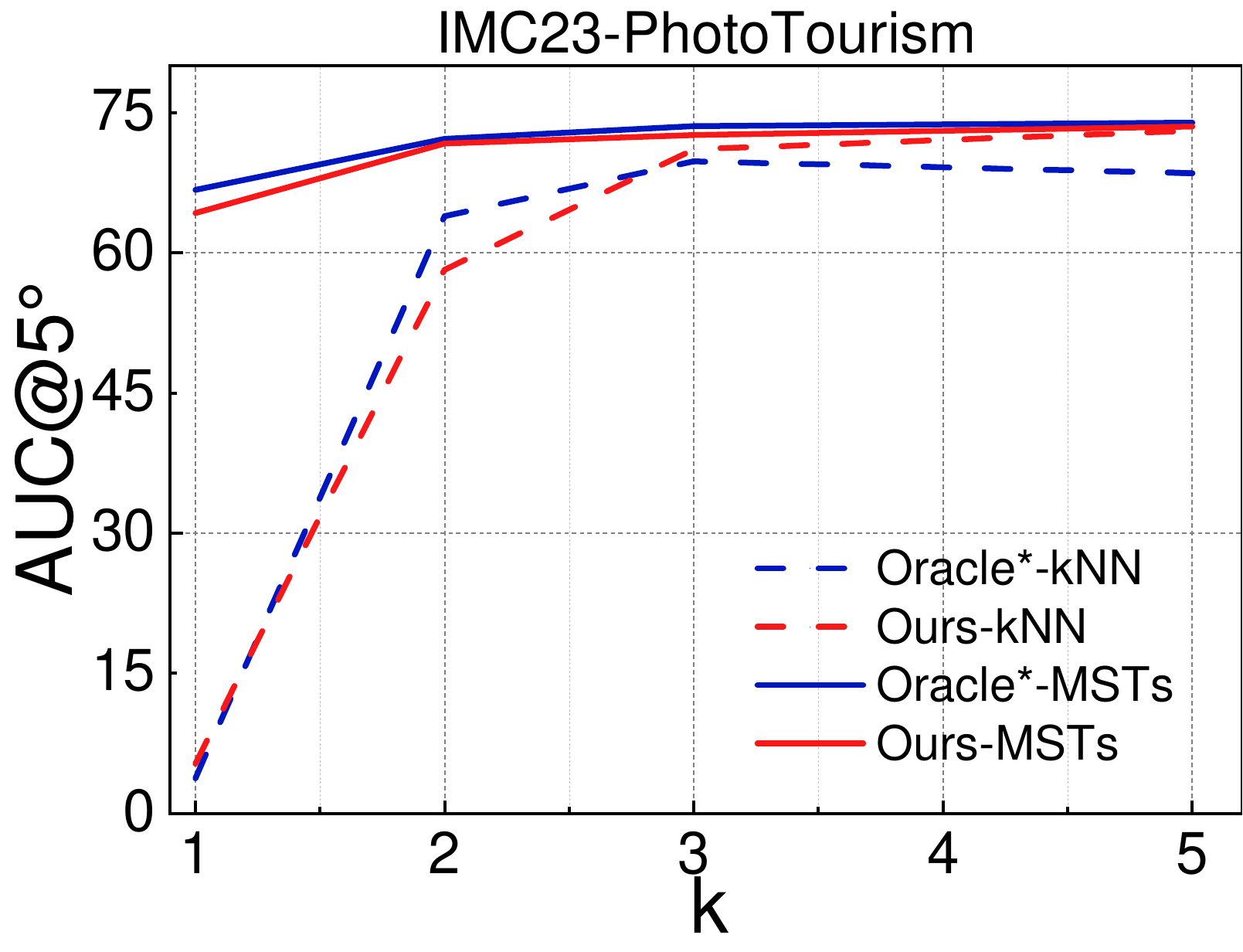}
   \includegraphics[width=0.33\textwidth, trim=5 0 0 0, clip]{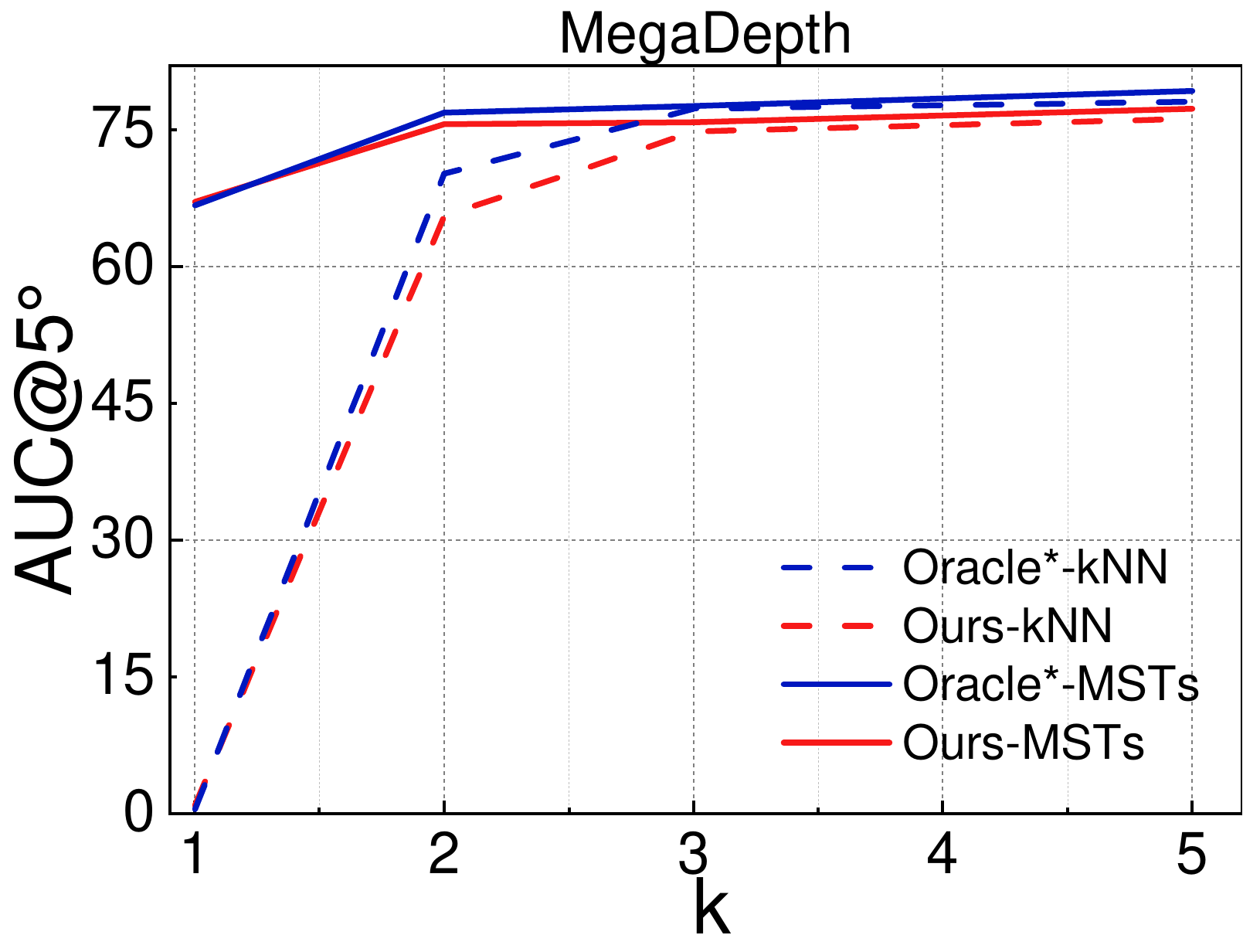}
      \includegraphics[width=0.33\textwidth, trim=5 0 0 0, clip]{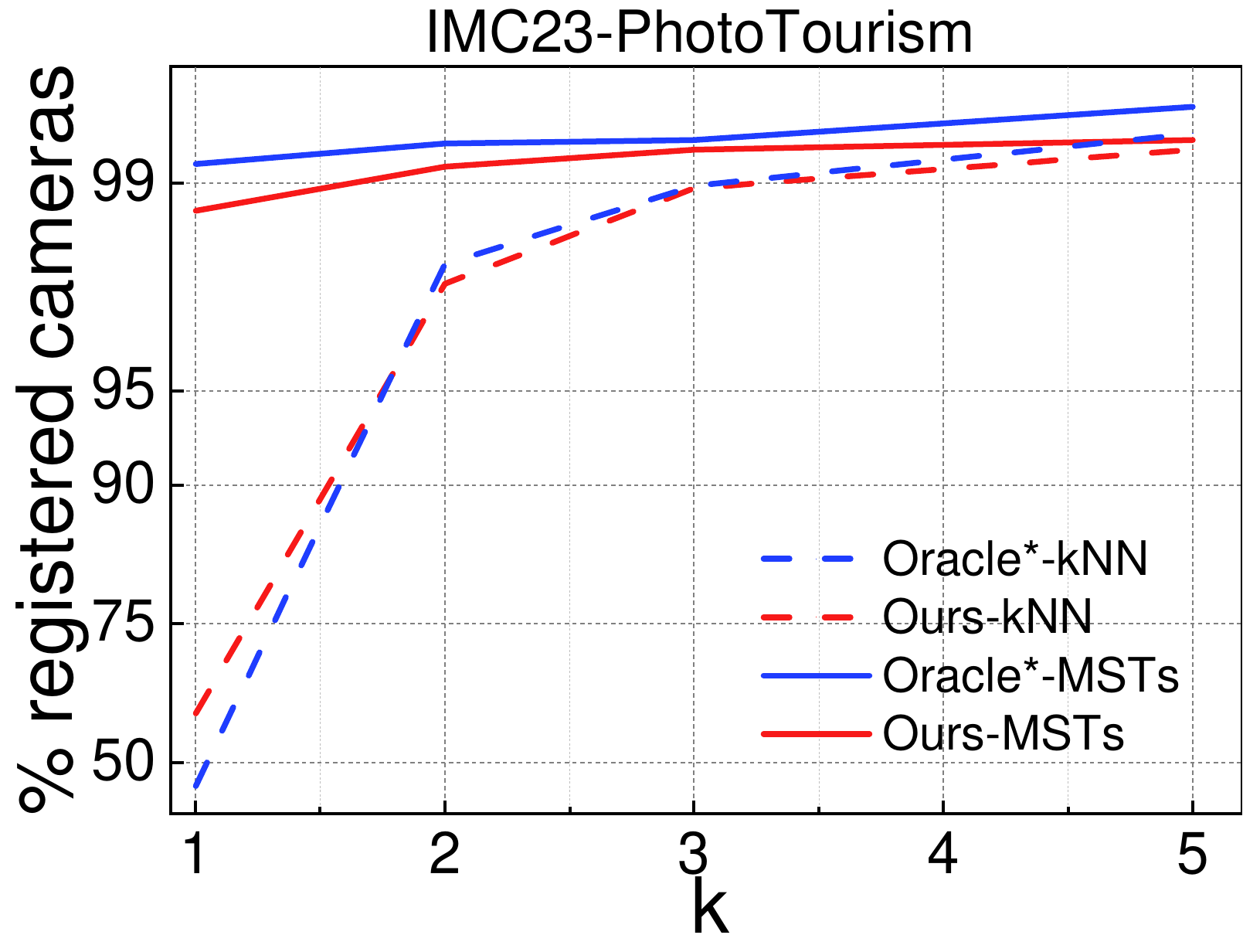}
    \vspace{-2mm}
    \caption{
        \textbf{Edge selection strategies} for pose graph initialization.  
        We report COLMAP~\cite{schonberger2016structure} relative pose AUC@5$^\circ$ on IMC23-PhotoTourism~\cite{imc2023} (\textit{left}) and MegaDepth~\cite{li2018megadepth} (\textit{middle}) when using either $k$NN or the proposed $k$ minimal spanning tree approach (MSTs) for graph construction.  
        Oracle$^\ast$ curves correspond to ground-truth rankings from SfM-derived geometry.  
        The percentage of successfully registered cameras on PhotoTourism is shown on the right.  
        MST-based selection yields substantially stronger connectivity and higher accuracy than $k$NN, and closely follows the oracle behavior across all $k$s.
    }
        
    \label{fig:knn}
    \vspace{0mm}
\end{figure*}
\boldparagraph{Edge Selection.}
We compare our multi-MST edge selection strategy against the standard $k$NN approach in Figure~\ref{fig:knn}.  
Using only the first nearest neighbor results in poor connectivity and consequently low pose accuracy, whereas a single MST guarantees a connected graph.  
As more MSTs are added, complementary long-range edges significantly improve accuracy while requiring far fewer candidate pairs than large-$k$ $k$NN.  
The right plot shows that MST-based selection yields substantially more complete reconstructions than $k$NN, which often fragments into disconnected components.

Oracle results -- using RANSAC inliers or common 3D points -- further show that MST-based selection closely aligns with geometric ground truth, producing more accurate and compact pose graphs than $k$NN.

\boldparagraph{Connectivity-Aware Score Modulation.}
Table~\ref{tab:mstupdates} reports ablations on VisymScenes.  
We disentangle:  
(i) whether modulation is applied,  
(ii) whether only the top-5 candidate edges per image are updated after each MST (to prevent unreliable edges from being boosted), and  
(iii) whether distances are normalized before modulation.

Modulation alone improves performance, especially at $k=2$, where early long-range edges are most influential.  
Restricting updates to top-5 candidates avoids strengthening weak edges and further improves accuracy.  
Distance normalization stabilizes the effect of modulation across scenes with different diameters.  
The best results arise when all components are enabled.

\begin{table}[t]
\centering
\hfill
\centering
\resizebox{0.45\textwidth}{!}{%
\begin{tabular}{lcccc}
\hline
 Variant&\multicolumn{4}{c}{AUC@5$^\circ~\uparrow$} \\
 $k$ MSTs~$\rightarrow$ & 1 & 2 & 3&5 \\ \hline
 Ours w/ SALAD backbone          & 61.2 & 71.0 &  72.4    & 73.4     \\
 Ours w/o GNN      & 55.4 & 70.4 & 72.3 &72.3 \\
 Ours                 & 64.2 & 71.7&72.6&73.5\\
 \hline
Oracle-RANSAC inliers & 65.7&72.4&73.0&74.1\\
Oracle-3D inliers & 65.4 & 72.1 & 73.5 & 74.3\\
 Oracle* &66.7&72.3 & 73.6 & 74.2 \\\hline
\end{tabular}%
}
\label{tab:ab}
\caption{
Ablation study on PhotoTourism evaluating different components of our method. 
We evaluate by using the weaker SALAD backbone and the removal of the GNN. 
The last three rows report oracle rankings derived directly from geometric signals (RANSAC inliers, 3D point overlap, or their combination), illustrating the relative strength of each supervision source. 
AUC@5$^\circ$ scores reported.
}
\label{tab:ab}
\end{table}

\boldparagraph{Training Components.}
Table~\ref{tab:ab} evaluates the impact of backbone choice, GNN reasoning, and supervision sources.  
Using a SALAD backbone slightly reduces accuracy, but the results remain well above those of raw SALAD features, showing that our method is backbone-agnostic.  
Removing the GNN leads to a large drop at $k=1$, underscoring the importance of global reasoning under sparse connectivity.

The last three rows compare oracle rankings based on RANSAC inliers, 3D inliers, and their combination.  
RANSAC inliers are most effective at small $k$, while 3D inliers excel at larger $k$.  
The combined supervision (Oracle*) offers the most stable performance.

\begin{table}[t]
\vspace{-3mm}
\centering

\centering
\resizebox{0.45\textwidth}{!}{%
\begin{tabular}{ccccccc}
\hline
 \multirow{2}{*}{Modul.} & Top-5 &Normalized &  \multicolumn{3}{c}{AUC@5$^\circ~\uparrow$}  \\
 & candidates &dist.& $k~\rightarrow$ 2 &  3  & 5 \\
\hline
 $\times$ &  $\times$ &$\times$ &61.9 &65.8 &\underline{71.7}\\
  $\times$  &  $\checkmark$ &$\times$ &61.4 &66.9 &64.8 \\
$\checkmark$  &  $\times$ &$\times$ &63.5 &64.7&68.9 \\
$\checkmark$  &  $\checkmark$ &$\times$ &63.7&65.5&68.3\\
$\times$  &  $\checkmark$ &$\checkmark$ &60.3& \underline{67.3}& 67.1 \\
$\checkmark$  &  $\times$ &$\checkmark$ &\bf{66.6} &66.3&71.3 \\
$\checkmark$  &  $\checkmark$ &$\checkmark$ & \underline{66.0} &\bf{71.5}&\bf{75.6} \\
\hline
\end{tabular}%
}
\vspace{0em}
\caption{
Ablation of the proposed connectivity-aware score modulation on VisymScenes.  
We evaluate three components: (i) whether modulation is applied, (ii) whether only the top-5 candidate edges per image are updated after each MST, and (iii) whether graph distances are normalized before modulation.  
Reported metrics are AUC@5$^\circ$ for pose graph initialization using $k\!=\!2,3,5$ MSTs.  
The best configuration combines modulation with normalized distances, yielding the strongest performance across all settings.}

\label{tab:mstupdates}

\end{table}

\vspace{-1mm}
\section{Conclusion}
\vspace{-1mm}

We introduced a framework for robust and efficient pose graph initialization based on \emph{global edge prioritization}. Our method predicts globally consistent matchability scores using a visual encoder and GNN-based message passing with geometry-derived supervision, addressing key limitations of retrieval-based initialization. Guided by these scores, we construct sparse yet well-connected pose graphs through a multi-minimal-spanning-tree strategy and further strengthen global connectivity via a distance-aware score modulation mechanism. Across PhotoTourism, MegaDepth, and VisymScenes, our approach consistently improves reconstruction accuracy, with the largest gains in the extremely sparse regime where long-range edges are most critical. Notably, on VisymScenes our method outperforms Doppelganger++ -- a dedicated doppelganger-filtering algorithm -- despite operating \emph{before} geometric verification, demonstrating strong robustness to severe visual ambiguities. These results show that integrating global reasoning directly into pose graph construction is a powerful direction for building faster and more reliable SfM pipelines. 


\section*{Acknowledgment}
Tong Wei was funded by Czech Technical University in
Prague grant No.~SGS23/173/OHK3/3T/13 and Research Center for Informatics (project CZ.02.1.01/0.0/0.0/16 019/0000765 funded by OP VVV).
We appreciate the constructive comments from all the reviewers and Dr. Paul‑Edouard Sarlin.

\section*{Appendix}
\appendix
\section{KNN-based Image Pair Selection}
Except for running COLMAP on MST pairs, We measure AUC@5$^\circ$ on IMC23 for MegaLoc~\cite{berton2025megaloc} and our method using $k$NN-based image-pair selection. As shown in the table below, our method outperforms the pre-trained model by a large margin on most values of $k$, with an exception at $k=2$. 
\begin{table}[h]
     \vspace{-3mm}
    \centering
 \resizebox{0.65\columnwidth}{!}{
    \begin{tabular}{c|c c c c}
    \toprule
       Method / $k$NNs  & k=1 & k=2 & k=3 & k=5    \\
       \midrule
       MegaLoc & 1.2 &\bf{63.9}  &64.0 & 66.5\\
       Ours &\bf{5.3} &  61.2  &\bf{71.1} & \bf{73.1}\\
        \bottomrule
    \end{tabular}}
    \vspace{-4mm}
\end{table}

\section{Graph Clustering}
Graph clustering is applied on large scale image collections based on the intermediate tokens output from our trained encoder.
After it is applied, the matchability scores are predicted by GNN and averaged on the overlapping edges among different subgraphs.
In our setup, the clustering is only applied on scenes with more than $500$ images. 
We tried applying the clustering approach applied to all scenes (including small ones), leading to competitive AUCs.

\section{Failure Cases}

In the figure below, failure examples of MegaLoc (unregistered cameras),  registered \textit{successfully} by our model, are shown in the first row.
In the second row, there are failure cases of both MegaLoc and ours (low resolution, small landmark part shown).

\begin{figure}[h]
    \centering
        \vspace{-3mm}
\includegraphics[width=0.14\linewidth]{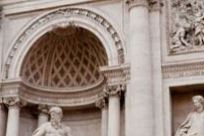}
\includegraphics[width=0.14\linewidth]{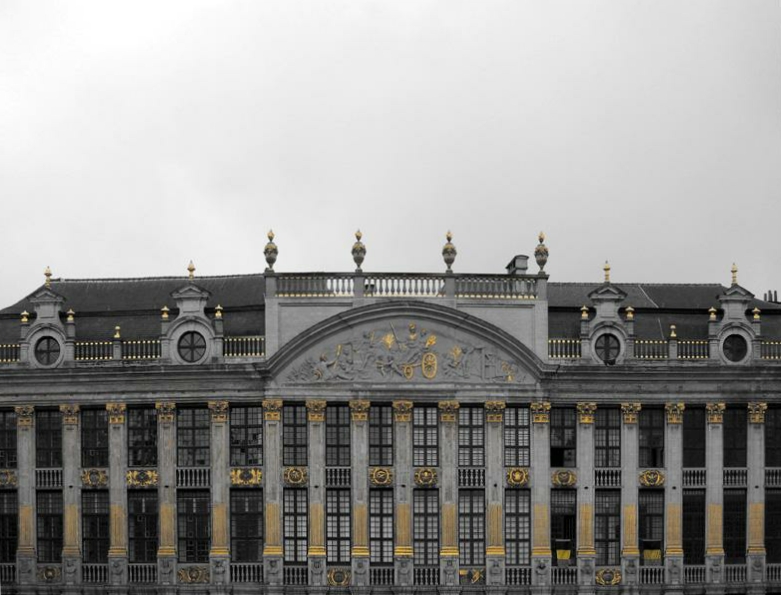}
\includegraphics[width=0.14\linewidth]{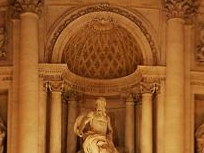}
\includegraphics[width=0.14\linewidth]{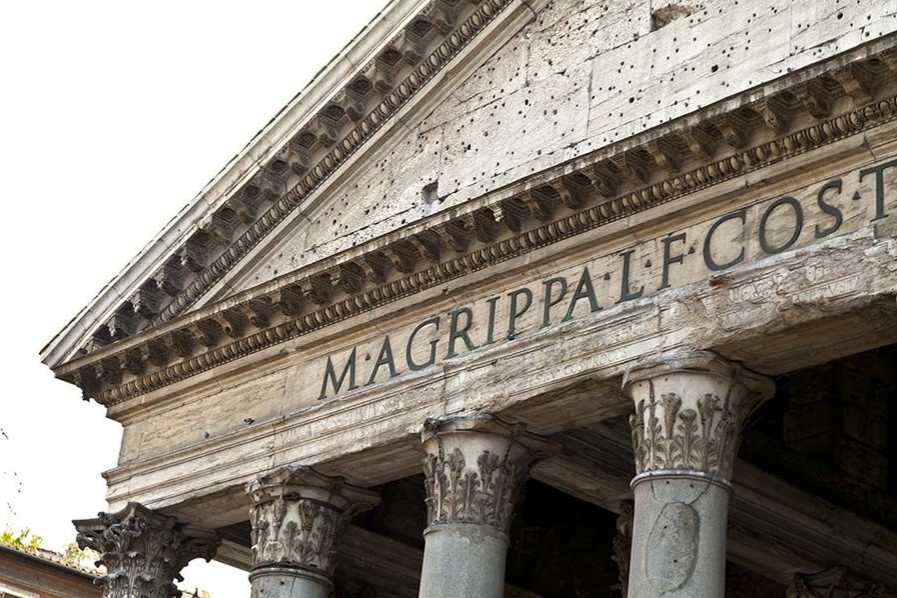}
\includegraphics[width=0.14\linewidth]{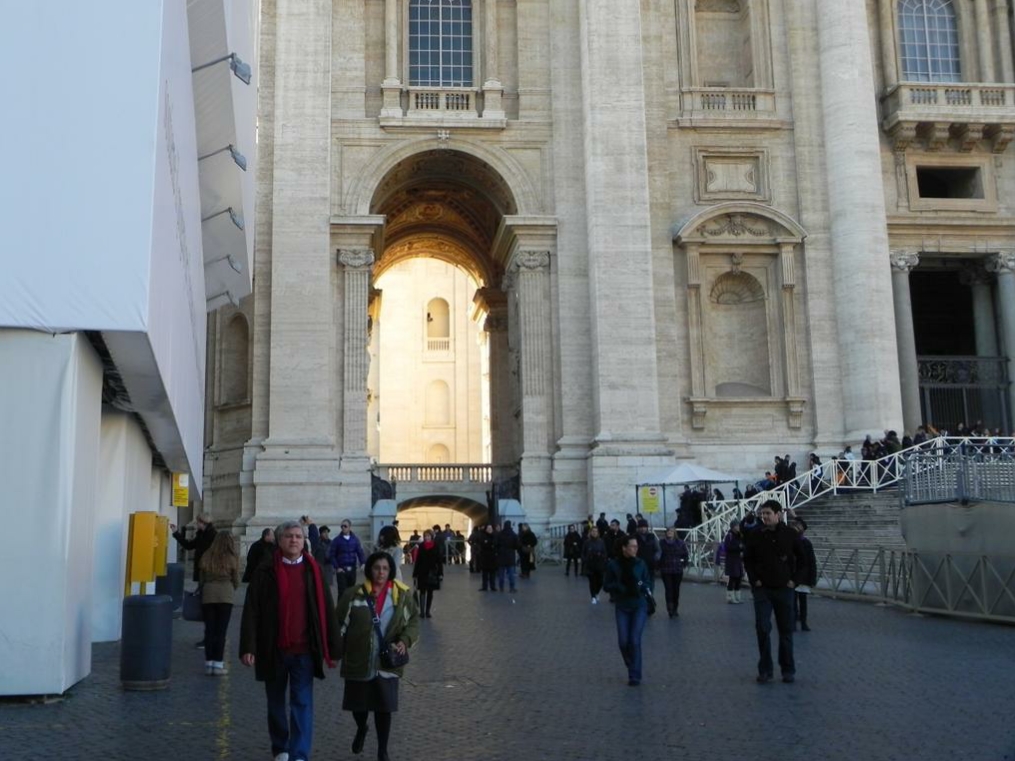}
\includegraphics[width=0.14\linewidth]{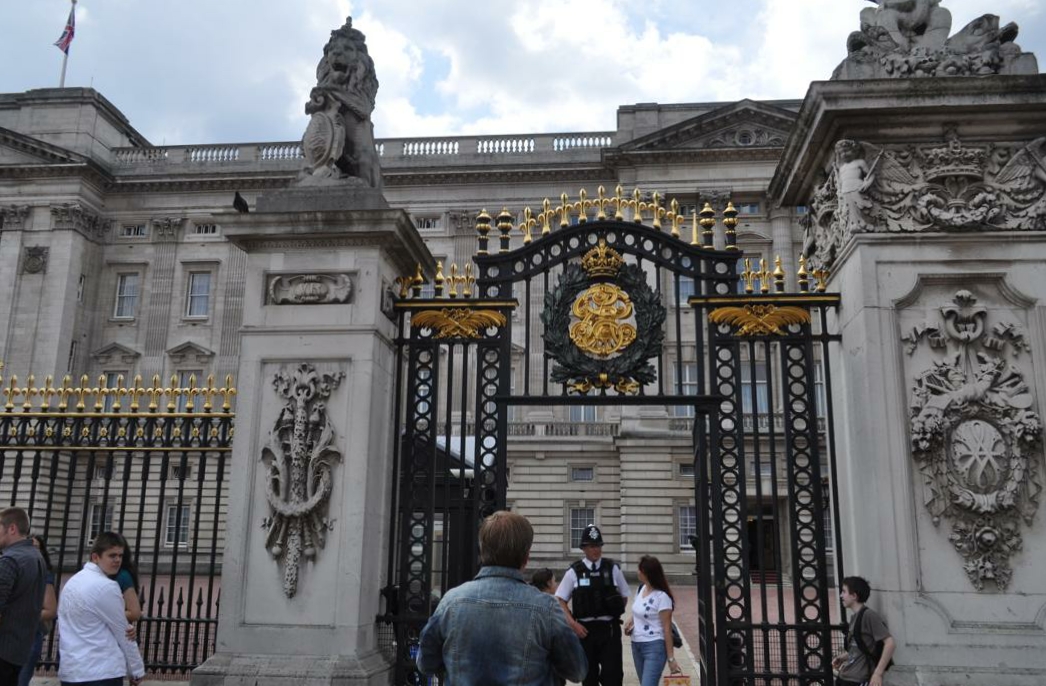}
\\
\includegraphics[width=0.14\linewidth]{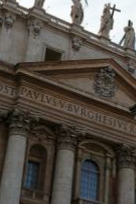}
\includegraphics[width=0.14\linewidth]{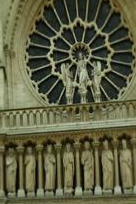}
\includegraphics[width=0.14\linewidth]{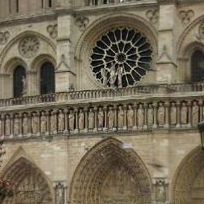}
\includegraphics[width=0.14\linewidth]{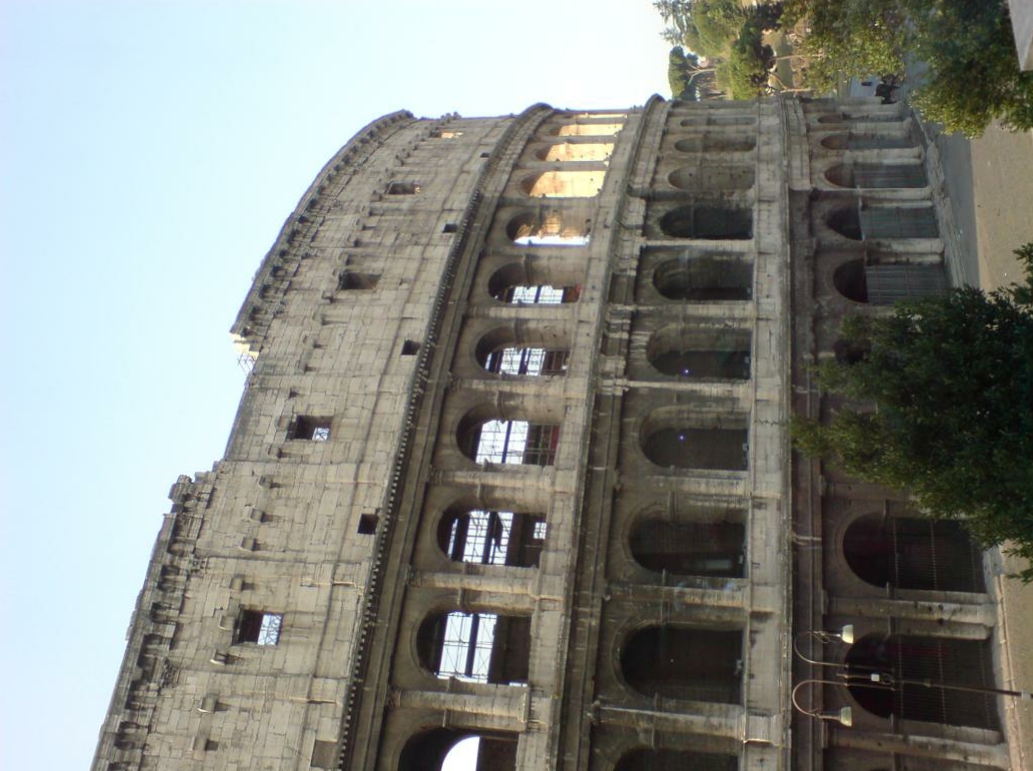}
\includegraphics[width=0.14\linewidth]{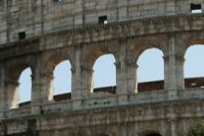}
\includegraphics[width=0.14\linewidth]{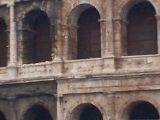}
    \vspace{-4mm}
\end{figure}

\section{Full Run-time Comparison}
The table below reports the timing~(\textit{in seconds}) for all steps of Ours and of MegaLoc, averaged on IMC23. 
Image pairs selected by two MSTs are fed in SfM.
Edge prediction for ours is slower than MegaLoc due to GNN, but this step takes negligible time compared to COLMAP. 
Note that COLMAP typically runs faster for ours due to better pair selection. 
\begin{table}[h]
     \vspace{-3mm}
    \centering
 \resizebox{0.75\columnwidth}{!}{
    \begin{tabular}{c|c c c c}
    \toprule
       Method   &Encoder & Predictor & MST &COLMAP   \\
       \midrule
       MegaLoc &\multirow{2}{*}{2.91} &\bf{0.01}&\bf{0.25} &2.3k\\
       Ours& &0.08&0.30 &\bf{2.1k}\\
        \bottomrule
    \end{tabular}}
    \vspace{-4mm}
\end{table}

\section{Backbone Discussion}
In the main paper, we have shown the performance of the model trained with MegaLoc and SALAD. 
To show how the model benefits from GNN predictor, we performed an experiment using DINOv2 (finetuning last 4 layers) and our GNN edge matchability predictor on top.
We observe that the GNN improves AUC@5$^\circ$ on IMC23 compared to the pretrained DINOv2 model.
\begin{table}[h]
     \vspace{-2mm}
    \centering
 \resizebox{0.68\columnwidth}{!}{
    \begin{tabular}{c|c c c c}
    \toprule
       Method / $k$-MSTs  & k=1 & k=2 & k=3 & k=5    \\
       \midrule
       pretrained DINOv2 &56.0& 67.4 & 70.1&72.4\\
       DINOv2 + GNN & \textbf{60.4} & \textbf{68.8} & \textbf{72.5} & \textbf{73.5}\\
        \bottomrule
    \end{tabular}}
    \vspace{-2mm}
\end{table}

{
    \small
    \bibliographystyle{ieeenat_fullname}
    \bibliography{main}
}

\end{document}